\documentclass[10pt,twocolumn,letterpaper]{article}

\usepackage{iccv}
\usepackage{times}

\usepackage[table,xcdraw]{xcolor}
\usepackage{cite}							
\usepackage[pdftex]{graphicx}				
\makeatletter
\@namedef{ver@everyshi.sty}{}
\makeatother
\usepackage{tikz}

\DeclareGraphicsExtensions{.pdf,.jpeg,.png}

\usepackage{tabularx}
\usepackage{amsmath}						
\usepackage{amssymb}
\usepackage{stmaryrd}
\usepackage{mathtools}
\usepackage{ifthen}
\usepackage{stix}
\usepackage{diagbox}
\usepackage{adjustbox}
\usepackage{algorithm}
\usepackage[noend]{algorithmic}

\usepackage{subcaption}

\usepackage{makecell}
\usepackage{array}   						
\usepackage{arydshln}
\usepackage{url}							
\usepackage{dsfont}
\usepackage{enumitem}

\usepackage[prependcaption,textsize=tiny]{todonotes}

\usepackage{pifont}

\usepackage[pagebackref,breaklinks,colorlinks]{hyperref}

\usepackage[capitalize]{cleveref}
\crefname{section}{Sec.}{Secs.}
\Crefname{section}{Section}{Sections}
\Crefname{table}{Table}{Tables}
\crefname{table}{Tab.}{Tabs.}
\crefname{algorithm}{Alg.}{Algs.}

\renewcommand\vec[1]{\boldsymbol{#1}}

\newcommand\R{\mathbb{R}}

\newcommand\X{\vec{X}}

\newcommand\x{\vec{x}}
\newcommand\y{\vec{y}}
\newcommand\z{\vec{z}}
\newcommand\p{\vec{p}}

\newcommand\tfinv{\mathcal{T}}

\newcommand\loss[1][]{%
	\ifthenelse { \equal {#1} {} }%
	{\mathcal{L}}%
	{\mathcal{L}_{\text{#1}}}%
}
\newcommand\distrib[1]{\mathbb{P}_{\text{#1}}}
\newcommand\dataset[1]{\mathcal{X}_{\text{#1}}}

\newcommand\ind[1]{\mathds{1}_{[#1]}}
\newcommand\cosim{\operatorname{sim}}

\newcommand\vA[1]{{#1}}
\newcommand\vB[1]{{#1'}}
\def\ascore{{s_a}}

\def\posA{{i}}
\def\posB{{j}}
\def\scaleK{{s}}
\def\batchK{{k}}

\newcommand\lnxent{\ell_{\text{NTX}}}

\def\softmax{{\operatorname{softmax}}}
\def\mem{{\operatorname{Mem}}}
\def\hopfield{{\operatorname{HF}}}
\def\var{{\operatorname{Var}}}
\DeclarePairedDelimiter\floor{\lfloor}{\rfloor}

\def\modelName{AnoMem}

\newcommand{\tablebottom}{\noalign{\hrule height 0.35ex}}
\newcommand{\tabletop}{\noalign{\hrule height 0.35ex}} %

\newcommand{\metricrel}[1]{\ \small{(\textcolor[rgb]{0,0.502,0}{#1})}}

\usepackage{layouts}

\iccvfinalcopy 

\def\iccvPaperID{****} 

\ificcvfinal\fi

\begin{document}

\title{Anomaly Detection via Multi-Scale Contrasted Memory}

\author{Loïc~Jézéquel${}^\star {}^\dagger$\hspace{15pt}
	Ngoc-Son~Vu$^\star$\hspace{15pt}
	Jean~Beaudet$^\dagger$\hspace{15pt} 
	Aymeric~Histace$^\star$ \vspace{5pt}
	\\ {\small ${}^\star$ ETIS UMR 8051 (CY Cergy Paris Université, ENSEA, CNRS) F-95000}
	\\ {\small ${}^\dagger$ Idemia Identity \& Security, 95520 Osny France}
}

\maketitle

\begin{abstract}
Deep anomaly detection (AD) aims to provide robust and efficient classifiers for one-class and unbalanced settings. However current AD models still struggle on edge-case normal samples and are often unable to keep high performance over different scales of anomalies. Moreover, there currently does not exist a unified framework efficiently covering both one-class and unbalanced learnings.
In the light of these limitations, we introduce a new two-stage anomaly detector which memorizes during training multi-scale normal prototypes to compute an anomaly deviation score. First, we simultaneously learn representations and memory modules on multiple scales using a novel memory-augmented contrastive learning. Then, we train an anomaly distance detector on the spatial deviation maps between prototypes and observations. 
Our model highly improves the state-of-the-art performance on a wide range of object, style and local anomalies with up to 50\% error relative improvement on CIFAR-100. It is also the first model to keep high performance across the one-class and unbalanced settings.
\end{abstract}

\vspace{-0.5cm}
\section{Introduction}
\label{sec:intro}

Detecting observations straying apart from a well defined normal baseline consistently lies at the center of many modern machine learning challenges. Given the complexity of the anomalous class and the high cost of obtaining labeled anomalies, this task of anomaly detection differs quite a lot from classical binary classification. This accordingly gave birth to many deep anomaly detection (AD) methods producing more stable results given an extremely unbalanced training dataset. Deep AD has been successful in various applications such as in fraud detection \cite{DBLP:conf/cvpr/FeiDYSX022}, medical imaging \cite{DBLP:conf/cvpr/SalehiSBRR21}, video surveillance \cite{DBLP:conf/cvpr/DoshiY21} or defect detection \cite{DBLP:conf/cvpr/RothPZSBG22}.

However, existing anomaly detection models still present some limitations. \textbf{(1)} There is a hard trade-off between remembering edge-case normal samples and remaining generalizable enough toward anomalies. This \textit{lack of normal sample memorization} often leads to high false reject rates on harder samples. \textbf{(2)} These models tend to focus on either local low-scale anomalies or global object oriented anomalies but fail to combine both. Current models often remain highly dataset-dependent and \textit{do not explicitly use multi-scaling}. \textbf{(3)} AD \textit{lacks an efficient unified framework} which could easily tackle one-class (OC) and semi-supervised (SS) detection. Indeed existing methods are either introduced as one-class or semi-supervised detectors, with different specialized approaches and set of hyper-parameters.

In the light of these limitations, we introduce in this paper a novel two-stage AD model named \modelName{} which memorizes during training multi-scale normal class prototypes to compute an anomaly deviation score at several scales. Unlike previous memory bank equipped methods \cite{DBLP:conf/iccv/GongLLSMVH19,DBLP:conf/cvpr/ParkNH20}, our normal memory layers cover the normal class at multiple scales and not only improve anomaly detection but also the quality of the learned representations. Additionally, by using the modern Hopfield layers for memorization, our method is much more efficient than nearest neighbor anomaly detectors \cite{DBLP:journals/corr/abs-2002-10445,DBLP:journals/corr/abs-2112-02597}. These detectors require keeping the whole normal set, while ours can learn the most representative samples with a fixed size. Through extensive experiments, we demonstrate that our method significantly outperforms all previous anomaly detectors using memory.

\begin{figure}[tb]
	\centering
	\resizebox{0.83\columnwidth}{!}{
		\begin{adjustbox}{clip,trim=0.3cm 0.30cm 0.2cm 0.25cm}
			\input{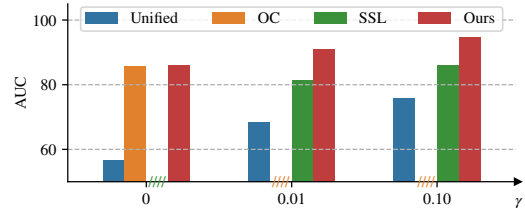}
		\end{adjustbox}
	}\vspace{-7pt}
	\caption{Comparison of our model with the best state-of-art methods on CIFAR-100 for different anomaly ratio $\gamma$.}
	\label{fig:perf-barplot}
\end{figure}

Our main contributions in this paper are the following:
\begin{itemize}[leftmargin=10pt,topsep=0pt,itemsep=0pt,partopsep=0pt,parsep=0pt]
\item We propose to integrate memory modules into contrastive learning to remember normal class prototypes during training. In the first stage, we simultaneously learn representations using contrastive learning and memory modules, allowing for effective \textbf{normal sample memorization}. In the second stage, we learn to detect anomalies using the memorized prototypes. When additional anomalous samples are available, we train an anomaly distance detector on the spatial deviation maps between prototypes and observations. To our best knowledge, \modelName{} is the first working well in both OC and SS settings with a few anomalies, making it \textbf{unifying method}.

\item \modelName{} is further improved by using multi-scale normal prototypes in both representation learning and AD stage. We introduce a novel way to efficiently memorize 2D features maps spatially. This enables our model to accurately detect low-scale, texture-oriented anomalies and higher-scale, object-oriented anomalies (\textbf{multi-scale AD}).

\item We validate the efficiency of our method and compare it with SoTA methods on one-vs-all, out-of-distribution (OoD) and face anti-spoofing detection. We improve detection with up to 50\% error relative improvement on object anomalies and 14\% on face anti-spoofing.
\end{itemize}

\section{Related work}

\begin{figure*}[tb]
	\centering
	\begin{subfigure}[b]{0.51\textwidth}
		\centering
		\includegraphics[trim=0 0 0.8cm 0,clip,width=1\columnwidth]{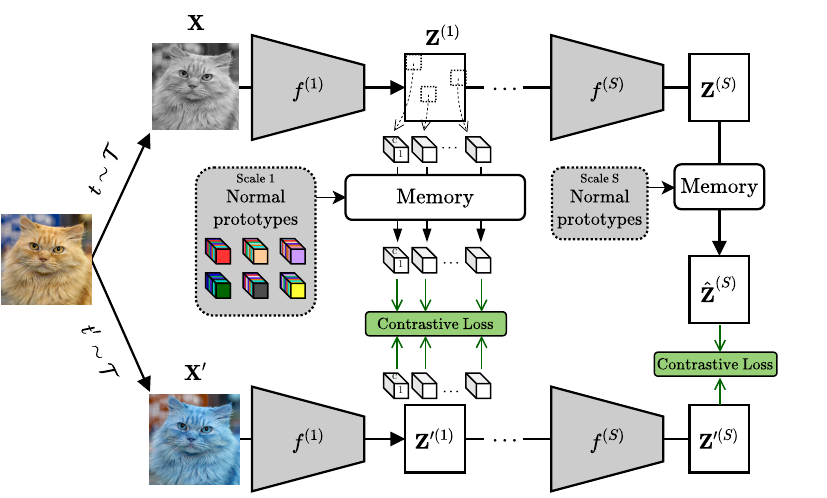}
		\vspace{-3.5pt}
        \caption{}
		\label{fig:model-train-stage1}
	\end{subfigure}%
	~ 
	\begin{subfigure}[b]{0.45\textwidth}
		\centering
		\includegraphics[width=1\columnwidth]{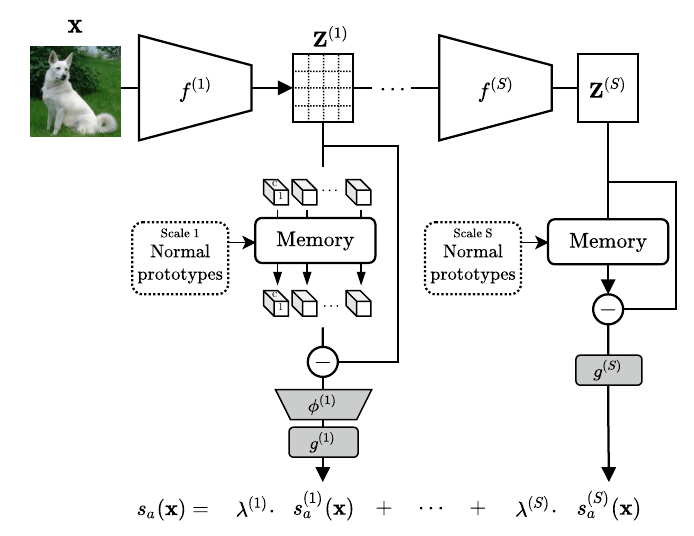}
        \caption{}
		\label{fig:model-train-stage2}
	\end{subfigure}
    \vspace{-5pt}
	\caption{Overview of \modelName{}'s training. (a) First representation learning stage where normal prototypes are learned at multiple scales using Hopfield layers and contrastive learning (b) Second anomaly detection stage where multi-scale anomaly distance models are learned from memory deviation maps. Learnable parts of the model are in dark gray.}
	\label{fig:model-train}
	\vspace{-9pt}
\end{figure*}

\subsection{Memory modules}
\label{sec:rel-memory}

A memory module should achieve two main operations: (i) writing inside a memory from a given set of samples (remembering) and (ii) recovering from a partial or corrupted input the most similar sample in its memory with minimal error (recalling). Most of the time, memory modules will differ on the amount of images they can memorize given a model size and the average reconstruction error.

A simple memory module is the nearest neighbor queue. Given a maximum size $M$, it stores the last $M$ samples representations in the queue. To remember an incomplete input $\x$, it retrieves the nearest neighbor from the queue.

A more effective memory module is the modern Hopfield layer \cite{DBLP:conf/iclr/RamsauerSLSWGHA21}. It represents the memory as a learnable matrix of weights $\X\in\R^{d\times N_{\text{Mem}}}$ and retrieves samples by recursively applying the following formula until convergence:
\begin{align}
	\vec{\xi}^{(t+1)}=\softmax{\left(\beta \vec{\xi}^{(t)}\X\right)}\X^T
\end{align} where $\vec{\xi}^{(0)}$ is the query vector and $\beta$ is the inverse temperature.
Its form is similar to the attention mechanism in transformers, except it reapplies the self-attention until convergence. This layer has a very high memory capacity and remember samples with very low redundancy \cite{DBLP:conf/iclr/RamsauerSLSWGHA21}. Subsequently, we call this layer a Hopfield layer of size $N_{\text{Mem}}$. 

\subsection{Anomaly detection (AD)}

AD is a binary classification problem where the normal class is usually well-defined as being sampled from a distribution $\distrib{norm}$, whereas the anomaly class is implicitly defined as anything not normal. The anomaly class is significantly broader and more complex than the normal class, creating a natural imbalance between the required amount of normal data and anomalous data. We call semi-supervised AD (\textbf{SSAD}) the setting where a small additional set of anomalies $\dataset{anom}$ is available. In one-class AD (\textbf{OC-AD}), only normal samples $\dataset{norm}$ are used for training.

There exist several families of approaches for OC-AD. \textbf{Pretext task methods} learn to solve on the normal data an auxiliary task \cite{DBLP:conf/avss/JezequelVBH21, DBLP:conf/nips/HendrycksMKS19,DBLP:conf/icpr/Jezequel22LPT, DBLP:conf/iclr/BergmanH20}, different from AD task. The inferred anomaly score describes how well the auxiliary task is performed on the input. Similarly, \textbf{two-stage methods} consist of a representation learning step and an anomaly score estimation step. After learning an encoder on normal data using self-supervised learning \cite{DBLP:conf/ijcai/ChoSL21,DBLP:conf/nips/TackMJS20,DBLP:conf/ijcai/ChenXLQZTZM21,DBLP:conf/icml/ZbontarJMLD21} or an encoder pre-trained on massive additional datasets \cite{DBLP:journals/corr/abs-2105-09270,DBLP:journals/corr/abs-2106-03844,DBLP:conf/cvpr/ReissCBH21}, the anomaly score is computed with a simple OC classifier on the latent space \cite{DBLP:conf/iclr/SehwagCM21, DBLP:conf/cvpr/LiSYP21, DBLP:conf/iclr/SohnLYJP21}. The methods \cite{DBLP:journals/corr/abs-2002-10445,DBLP:journals/corr/abs-2112-02597} have used a nearest neighbor queue to fetch the closest prototypical normal samples inside the latent space. The mean $L_2$ distance to these samples is then used as the anomaly score. \textbf{Density estimation methods} tackle the estimation of the distribution $\distrib{norm}$ by using deep high-dimensional density estimators such as normalizing-flows \cite{DBLP:journals/corr/abs-2106-12894}, likelihood ratio methods \cite{DBLP:conf/nips/RenLFSPDDL19}, variational models \cite{DBLP:journals/corr/abs-1911-04971} or more recently diffusion models \cite{DBLP:conf/cvpr/WyattLSW22, DBLP:journals/corr/abs-2205-14297}. \textbf{Reconstruction methods} measure the reconstruction error of a bottleneck encoder-decoder, trained using denoising autoencoders \cite{DBLP:conf/cvpr/PereraNX19, DBLP:conf/cvpr/SchneiderASS22} or two-way GANs \cite{DBLP:conf/accv/AkcayAB18,DBLP:conf/acpr/TuluptcevaBFK19,DBLP:conf/ipmi/SchleglSWSL17, DBLP:conf/icip/LiuLZHW21}. Some works used memory in the latent space of an auto-encoder \cite{DBLP:conf/iccv/GongLLSMVH19,DBLP:conf/cvpr/ParkNH20} for AD. During training the latent memory weights are learned to achieve optimal reconstruction on the normal class. Then, the reconstruction is performed from the memory closest fetched latent vector. More recently, \textbf{knowledge distillation methods} have been adapted to AD by using the representation discrepancy of anomalies in the teacher-student model \cite{DBLP:conf/cvpr/DengL22, DBLP:conf/cvpr/CohenA22}.

SSAD mainly revolves around two-stage methods and anomaly distance methods. We note however that some recent work has tried to generalize pretext task to SSAD \cite{DBLP:conf/icpr/Jezequel22LPT}. In the \textbf{SSAD two-stage methods}, a supervised classifier with the anomalous samples is trained in the second stage instead of the aforementioned one-class estimator \cite{DBLP:conf/bmvc/0001S0PC21}. \textbf{Distance methods} directly use a distance to a centroid as the anomaly score and learn the model to maximize the anomaly distance on anomalous samples and minimize it on normal samples \cite{DBLP:conf/iclr/RuffVGBMMK20, DBLP:conf/icpr/Jezequel22CR}. To the best of our knowledge, \textit{no SSAD methods in the literature use any kind of memory mechanism for the anomaly score computation}.

We also note that there is a closely related task of anomaly localization (AL), which goal is to produce an anomaly heatmap. AL datasets range from defect localization \cite{DBLP:conf/cvpr/BergmannFSS19} to surveillance video abnormal event detection \cite{DBLP:conf/cvpr/MahadevanLBV10}. Specialized methods targeting localization \cite{DBLP:conf/accv/YiY20, DBLP:conf/cvpr/LiSYP21} have been introduced to efficiently solve this task.

\subsection{Contrastive learning}

Contrastive learning is a self-supervised representation learning method. It operates on the basis of two principles: \textbf{(1)} two different images should yield dissimilar representations, and \textbf{(2)} two different views of the same image should yield similar representations. The views of an image are characterized by a set of transformations $\tfinv$. There have been many methods enforcing these two principles: SimCLR \cite{DBLP:conf/icml/ChenK0H20}, Barlow-Twins \cite{DBLP:conf/icml/ZbontarJMLD21} and VICReg \cite{DBLP:conf/iclr/BardesPL22} with a siamese network, MoCo \cite{DBLP:conf/cvpr/He0WXG20} with a negative sample bank, BYOL \cite{DBLP:conf/nips/GrillSATRBDPGAP20} and SimSiam \cite{DBLP:conf/cvpr/ChenH21} with a teacher-student network or SwAV \cite{DBLP:conf/nips/CaronMMGBJ20} with contrastive clusters. While some contrastive methods such as SimCLR and MoCo require negatives samples, other such as BYOL, SimSiam and SwAV do not.

In the simplest formulation, the pairs of representations to contrast are only considered from two views of a batch augmented by transformations $(\vA{t},\vB{t})$. In \textbf{SimCLR}, the following loss is minimized on those two batches: \vspace{-0.3cm}
\begin{align}
	\label{eq:simclr}
	\loss[CO]=\frac{1}{2B}\sum_{\batchK=1}^B \lnxent(\vA{\z}_\batchK,\vB{\z}_\batchK)+\lnxent(\vB{\z}_\batchK,\vA{\z}_\batchK)
\end{align} where $\vA{\z}, \vB{\z}$ are the last features of the two augmented batches and $\lnxent$ is the Normalized Temperature-scaled Cross Entropy Loss (\textbf{NT-Xent}). In practice, minimizing $\loss[CO]$ will yield representations with the most angular spread variance, while retaining angular invariance in regards to $\tfinv$. 

Memory mechanisms can also be used into contrastive learning as proposed in \cite{DBLP:conf/iccv/DwibediATSZ21, DBLP:conf/iccv/KoohpayeganiTP21}. During training, the positive and negative pairs are augmented with the samples nearest neighbors from a memory queue. This allows the method to reach better performance for smaller batch sizes.

\section{Proposed method}

\cref{sec:memorize} first details a novel training procedure to simultaneously learn an encoder representations and a set of multi-scale normal class prototypes. \cref{sec:normal-prot-dev} then presents how to use the encoder and the normal prototypes to train a one-class or unbalanced anomaly detector in a unified framework. Our model training is fully summarized in \cref{fig:model-train}.

\textbf{Notation.} Let $\dataset{}=\{(\x_\batchK,y_\batchK)\}_\batchK$ be a training dataset made of normal samples ($y_\batchK=1$) and potentially of anomalies ($y_\batchK=0$). We use a backbone network $f:\mathbb{R}^{H\times W\times C}\rightarrow \mathbb{R}^D$ composed of several stages $f^{(1)},\cdots,f^{(S)}$ such that the dimensions of the $\scaleK$\textsuperscript{th} scale feature map are $H^{(\scaleK)}\times W^{(\scaleK)}\times C^{(\scaleK)}$. We note $\z^{(\scaleK)}=f^{(\scaleK)}\circ\cdots\circ f^{(1)}(\x)$.

\subsection{Memorizing normal class prototypes}
\label{sec:memorize}

In this section, we first introduce how memory modules can be used in the contrastive learning scheme to provide robust and representative normal class prototypes. Then we generalize our idea to several scales throughout the encoder.

Foremost, we choose to apply a contrastive learning type method rather than other unsupervised learning schemes, as it produces better representations with very few labeled data \cite{DBLP:journals/corr/abs-2105-05837}. We also favor self-supervised learning on the normal data rather than using a pre-trained encoder on generic datasets \cite{DBLP:journals/corr/abs-2105-09270} which often performs poorly on data with a significant distribution shift. In order to learn unsupervised representations and a set of normal prototypes, we could sequentially apply contrastive learning then perform k-means clusterisation and use the cluster centroids as the normal prototypes. However this approach has two main flaws. First, the representation learning step and the construction of prototypes are completely separated. Indeed, it has been shown in several contrastive learning methods \cite{DBLP:conf/iccv/DwibediATSZ21, DBLP:conf/nips/CaronMMGBJ20} that the inclusion of a few representative samples in the negative examples can significantly improve the representation quality, and alleviate the need for large batches. Moreover, the resulting k-means prototypes do not often cover atypical samples. This means that harder normal samples will not be well encompassed by the normal prototypes, resulting in high false rejection rate during AD. We compare our approach with k-means centroids in \cref{sec:abl-study}.

In the light of the aforementioned pipeline flaws, we introduce a novel approach based on memory modules to jointly learn an encoder and normal prototypes. Let $\vA{\z}_\batchK$ and $\vB{\z}_\batchK$ respectively be the encoder features for the contrastive upper and lower branch. Instead of contrasting $\vA{\z}_\batchK$ and $\vB{\z}_\batchK$, we apply beforehand an Hopfield memory layer $\hopfield(\cdot)$ to the first branch in the case of normal samples. We note that the memory layer is only applied when the sample is normal since we assumed that anomalous data is significantly more variable and less defined than normal data. As such, we note \begin{equation}\label{eq:memLayer}
	\mem(\z, y)=y\cdot \hopfield(\z)+(1-y)\cdot\z
\end{equation}
We choose SimCLR as the contrastive loss baseline, but our method can be integrated into any other contrastive framework. Indeed, we could use Barlow-Twins by replacing $\lnxent$ with the cross-correlation. We then use the following contrasted memory loss for representation learning:
\vspace{-4pt}\begin{align}
	\loss[COM](\vA{\z},\vB{\z},y)=\frac{1}{2B}\sum_{\batchK=1}^B & 
	\left[ \lnxent\left(\mem(\vA{\z}_\batchK, y_\batchK),\vB{\z}_\batchK\right)+ \right. \nonumber\\ 
	&	\left.\ \lnxent\left(\vB{\z}_\batchK,\mem(\vA{\z}_\batchK, y_\batchK)\right) \right]
\end{align} where \begin{align}
\lnxent(\vA{\z}, \vB{\z})=-\log\frac{\exp(\cosim(\vA{\z},\vB{\z})/\tau)}{\sum_{\p}\ind{\p\neq\vA{\z}}\exp(\cosim(\vA{\z},\p)/\tau)}
\end{align} 
$\p$ covers any representation inside the multi-view batch, $\tau$ is a temperature hyper-parameter and $\cosim(\cdot,\cdot)$ is the cosine similarity. In contrast with existing two-stage AD methods, we explicitly introduces the anomalous and normal labels from the very first step of representation learning. 

\vspace{-6pt}\paragraph{Variance loss as regularization.} Our procedure can be prone to representation collapse during the first epochs. Indeed, we observed that the dynamic between contrastive learning and the randomly initialized memory layer can occasionally lead to a collapse of all prototypes to a single point during the first epochs. To prevent this, we introduce an additional regularization loss which ensures the variance of the retrieved memory samples does not reach zero:
\vspace{-4pt}\begin{align}
	\loss[V](\z,y)=-\frac{1}{\sum_\batchK y_\batchK}\sum_{\batchK=1}^B y_\batchK \sqrt{\var\left[\mem(\z_\batchK, y_\batchK)\right]}
\end{align}

\paragraph{Multi-scale contrasted memory.} To gather information from several scales, we apply our contrasted memory loss not only to the flattened 1D output $\z$ of our encoder but also to $S$ intermediate layer 3D feature maps $\z^{(1)},\cdots,\z^{(S)}$. 

We add after each scale representation $\z^{(\scaleK)}$ a memory layer $\hopfield^{(\scaleK)}$ to effectively capture multi-scale normal prototypes. However, memorizing the full 3D map as a single flattened vector would not be ideal. Indeed, at lower scales we are interested in memorizing local patterns regardless of their position. Moreover, the memory would span across a space of very high dimensions. Therefore, 3D intermediate maps are viewed as a collection of $H^{(\scaleK)} W^{(\scaleK)}$ 1D feature vectors $\z^{(\scaleK)}_{\posA,\posB}$ rather than a single flattened 1D vector. This is equivalent to remembering the image as patches. 

Since earlier features map will have a high resolution, the computational cost and memory usage of such approach can quickly explode. Thus, we only apply our contrasted memory loss on a random sample with ratio $r^{(\scaleK)}$ of the available vectors on the $\scaleK$\textsuperscript{th} scale.

Our \textbf{multi-scale contrasted memory loss} becomes
\vspace{-8pt}\begin{align}
	\label{eq:comms}
	\loss[{\scriptsize COM-MS}]&=\frac{1}{\sum_\scaleK |\Omega^{(\scaleK)}|}\sum_{\scaleK=1}^S\sum_{(\posA,\posB)\in \Omega^{(\scaleK)}} \lambda^{(\scaleK)} \left[\vphantom{\sum}\right.\nonumber\\ 
	&\left. \loss[COM]\left(\vA{\z}^{(\scaleK)}_{\posA,\posB}, \vB{\z}^{(\scaleK)}_{\posA,\posB}, y\right) +  \lambda_V\loss[V](\vA{\z}^{(\scaleK)}_{\posA,\posB},y)\right]
\end{align} where $\lambda_V$ controls the impact of the variance loss, $\lambda^{(\scaleK)}$ controls the importance of the $\scaleK$\textsuperscript{th} scale, and  $\Omega^{(\scaleK)}$ is a random sample without replacement of $\floor{H^{(\scaleK)}\  W^{(\scaleK)}\ r^{(\scaleK)}}$ points from $\llbracket 1,H^{(\scaleK)} \rrbracket\times \llbracket 1,W^{(\scaleK)} \rrbracket$. We choose to put more confidence on the latest stages which are more semantically meaningful than earlier scales, meaning that $\lambda^{(1)}<\cdots<\lambda^{(S)}$.

We simultaneously minimize this loss on all of the encoder stages and the $S$ memory layers' weights. An overview of this first stage is given in \cref{fig:model-train-stage1}, and its algorithm is presented in \cref{alg:step1}. Compared to previous memory bank equipped anomaly detectors \cite{DBLP:conf/iccv/GongLLSMVH19,DBLP:conf/cvpr/ParkNH20,DBLP:journals/corr/abs-2002-10445,DBLP:journals/corr/abs-2112-02597}, our model is the first to memorize the normal class at several scales allowing it to be more robust to anomaly sizes. Moreover, the use of normal memory does not only improve anomaly detection but also the quality of the learned representations, as will be discussed in \cref{sec:abl-study}. 

\begin{algorithm}[tb]
	\footnotesize
	\caption{\modelName{} first learning stage}\label{alg:step1}
	\begin{algorithmic}[1]
		\STATE {\bfseries Input:} batch size $B$, invariance transformations $\tfinv$
		\STATE {\bfseries Initialization:} encoders $f^{(1)},\cdots,f^{(S)}$, 
		memory $\text{HF}^{(1)},\cdots,\text{HF}^{(S)}$.
		\WHILE{not reach the maximum epoch} 
		\STATE Sample image minibatch  $\X$ with labels $\y$ 
		\STATE Sample augmentations $\vA{t},\vB{t}$ from $\tfinv$
		
		\STATE Get augmented views $\vA{\mathbf{Z}}^{(0)}\leftarrow \vA{t}(\mathbf{X})$ and $\vB{\mathbf{Z}}^{(0)}\leftarrow \vB{t}(\mathbf{X})$%
		
		\FOR{$\scaleK=1\cdots S$}
		\STATE 
		$\vA{\mathbf{Z}}^{(\scaleK)}\leftarrow f^{(\scaleK)}(\vA{\mathbf{Z}}^{(\scaleK-1)})$ and $\vB{\mathbf{Z}}^{(\scaleK)}\leftarrow f^{(\scaleK)}(\vB{\mathbf{Z}}^{(\scaleK-1)})$
		\STATE Sample $\lfloor H^{(\scaleK)}\ W^{(\scaleK)}\  r^{(\scaleK)}\rfloor$ vectors $\vA{Z}_{\posA,\posB}^{(\scaleK)}$ from $\vA{\mathbf{Z}}^{(\scaleK)}$
		\STATE Retrieve each $\vA{Z}_{\posA,\posB}^{(\scaleK)}$ memory prototypes using \cref{eq:memLayer} with the $\scaleK$\textsuperscript{th} scale memory layer.

		\ENDFOR
		\STATE Compute $\loss[COM-MS]$ from \cref{eq:comms}
		\STATE Gradient descent on $\loss[COM-MS]$ to update $f^{(1)},\cdots,f^{(S)}$
		 and $\text{HF}^{(1)},\cdots,\text{HF}^{(S)}$.
		\ENDWHILE

		\STATE {\bfseries Output:} {\bfseries Encoder network} $f$, and the {\bfseries multi-scale memory prototypes} from $\mem^{(1)},\cdots,\mem^{(S)}$.
	\end{algorithmic}
 \end{algorithm}

\subsection{Multi-scale normal prototype deviation}
\label{sec:normal-prot-dev}

 In this second step of training, our goal is to compute an anomaly score given the pre-trained encoder $f$ and the multi-scale normal memory layers $\hopfield^{(1)},\cdots,\hopfield^{(S)}$.

For each scale $\scaleK$, we consider the difference $\vec{\Delta}^{(\scaleK)}$ between the encoder feature map $\z^{(\scaleK)}$ and its recollection from the $\scaleK$\textsuperscript{th} memory layer. The recollection process consists in spatially applying the memory layer to each $C^{(\scaleK)}$ depth 1D vector:
\begin{align}
	\vec{\Delta}^{(\scaleK)}=\z^{(\scaleK)}-\hopfield^{(\scaleK)}\left(\z^{(\scaleK)}\right)
\end{align} where $(\hopfield(\z))_{\posA,\posB} = \hopfield(\z_{\posA,\posB})$. 

\textbf{One-class AD.} In this case, we use the $L_2$ norm of the difference map as an anomaly score for each scale and no further training is required:
\begin{align}
	\ascore^{(\scaleK)}(\x)=\|\Delta^{(\scaleK)}\|_2
\end{align}

\textbf{Semi-supervised AD.} We use the additional labeled data to train $S$ scale-specific classifiers on the difference map $\Delta^{(\scaleK)}$. Each classifier is first composed of an average pooling layer $\phi$ followed by a two-layer MLP $g^{(\scaleK)}$ with a single scalar output. $\phi$ reduces the spatial resolution of $\Delta^{(\scaleK)}$, to prevent using very large layers on earlier scales. The output of $g^{(\scaleK)}$ directly corresponds to the $\scaleK$\textsuperscript{th} scale anomaly distance: \vspace{-1pt}\begin{align}
	\ascore^{(\scaleK)}(\x)=g^{(\scaleK)}\circ\phi(\Delta^{(\scaleK)})
\end{align}

Each scale-specific classifier is trained using the intermediate features of the same normal and anomalous samples used during the first step along their labels. The training procedure is similar to other distance-based anomaly detectors \cite{DBLP:conf/iclr/RuffVGBMMK20, DBLP:conf/icpr/Jezequel22CR} where the objective is to obtain small distances for normal samples while keeping high distances on anomalies. We note that our model is the \textit{first to introduce memory prototypes learned during representation learning into the anomaly distance learning}. The distance constraint is enforced via a double-hinge distance loss:
\begin{align}
	\resizebox{0.88\columnwidth}{!}{%
	$\ell_{\text{dist}}(d, y)=y{\cdot}\max\left(d{-}\frac{1}{M}, 0\right){+}(1{-}y){\cdot}\max\left(M{-}d,0\right)$
	}
\end{align} where $d$ is the anomaly distance for a given sample, and $M$ controls the size of the margin around the unit ball frontier. Using this loss, both normal samples and anomalies will be correctly separated without encouraging anomalous features to be pushed toward infinity. Our \textbf{second stage supervised loss} is the following
\vspace{-3pt}\begin{align}
	\loss[SUP]=\frac{1}{S\cdot B}\sum_{\batchK=1}^B\sum_{\scaleK=1}^S \ell_{\text{dist}}\left(g^{(\scaleK)}\circ\phi(\Delta^{(\scaleK)}),y_\batchK\right)
\end{align}

Finally, all scale anomaly scores are merged using a sum weighted by the confidence parameters $\lambda^{(\scaleK)}$:
\begin{equation}
	\ascore(\x)=\frac{1}{\sum_\scaleK\lambda^{(\scaleK)}}\sum_\scaleK \lambda^{(\scaleK)}\cdot \ascore^{(\scaleK)}(\x)
\end{equation} The anomaly score $\ascore$ effectively combines the expertise of the scores from each scale, making it more robust to different sizes of anomalies than other detectors. As mentioned in \cref{sec:memorize}, the $\lambda^{(\scaleK)}$ will put more weight to later scales, making our detector more sensitive to broad object anomalies.

This second stage is summarized in \cref{fig:model-train-stage2}. As we can see, only the second stage has to be swapped between one-class learning and semi-supervised learning, resulting in a unified easily-switchable framework for AD.

\begin{table*}[t]
	\centering
	\resizebox{0.73\textwidth}{!}{%
		\begin{tabular}{l|cccc|cccc;{1pt/1pt}cccc} 
\tabletop\textbf{AUROC (\%)} 	 & 	 \multicolumn{4}{c|}{\textbf{CUB-200}} 	 & 	 \multicolumn{4}{c;{1pt/1pt}}{\textbf{CIFAR-10}} 	 & 	 \multicolumn{4}{c}{\textbf{CIFAR-100}} \\ 
\textbf{Models \textbackslash{} $\boldsymbol{\gamma}$} 	 & 	 \textbf{0.} 	 & 	 \textbf{0.01} 	 & 	 \textbf{0.05} 	 & 	 \textbf{0.10} 	 & 	 \textbf{0.} 	 & 	 \textbf{0.01} 	 & 	 \textbf{0.05} 	 & 	 \textbf{0.10} 	 & 	 \textbf{0.} 	 & 	 \textbf{0.01} 	 & 	 \textbf{0.05} 	 & 	 \textbf{0.10} \\ 
\hline 
MemAE \cite{DBLP:conf/iccv/GongLLSMVH19} 	 & 	 59.6 	 & 	 \cellcolor[HTML]{F5F5F5}{ } 	 & 	 \cellcolor[HTML]{F5F5F5}{ } 	 & 	 \cellcolor[HTML]{F5F5F5}{ } 	 & 	 60.9 	 & 	 \cellcolor[HTML]{F5F5F5}{ } 	 & 	 \cellcolor[HTML]{F5F5F5}{ } 	 & 	 \cellcolor[HTML]{F5F5F5}{ } 	 & 	 57.4 	 & 	 \cellcolor[HTML]{F5F5F5}{ } 	 & 	 \cellcolor[HTML]{F5F5F5}{ } 	 & 	 \cellcolor[HTML]{F5F5F5}{ } \\ 
OC-SVM \cite{DBLP:conf/nips/ScholkopfWSSP99} 	 & 	 76.3 	 & 	 \cellcolor[HTML]{F5F5F5}{ } 	 & 	 \cellcolor[HTML]{F5F5F5}{ } 	 & 	 \cellcolor[HTML]{F5F5F5}{ } 	 & 	 64.7 	 & 	 \cellcolor[HTML]{F5F5F5}{ } 	 & 	 \cellcolor[HTML]{F5F5F5}{ } 	 & 	 \cellcolor[HTML]{F5F5F5}{ } 	 & 	 62.6 	 & 	 \cellcolor[HTML]{F5F5F5}{ } 	 & 	 \cellcolor[HTML]{F5F5F5}{ } 	 & 	 \cellcolor[HTML]{F5F5F5}{ } \\ 
PIAD \cite{DBLP:conf/acpr/TuluptcevaBFK19} 	 & 	 63.5 	 & 	 \cellcolor[HTML]{F5F5F5}{ } 	 & 	 \cellcolor[HTML]{F5F5F5}{ } 	 & 	 \cellcolor[HTML]{F5F5F5}{ } 	 & 	 79.9 	 & 	 \cellcolor[HTML]{F5F5F5}{ } 	 & 	 \cellcolor[HTML]{F5F5F5}{ } 	 & 	 \cellcolor[HTML]{F5F5F5}{ } 	 & 	 78.8 	 & 	 \cellcolor[HTML]{F5F5F5}{ } 	 & 	 \cellcolor[HTML]{F5F5F5}{ } 	 & 	 \cellcolor[HTML]{F5F5F5}{ } \\ 
GOAD \cite{DBLP:conf/iclr/BergmanH20} 	 & 	 66.6 	 & 	 \cellcolor[HTML]{F5F5F5}{ } 	 & 	 \cellcolor[HTML]{F5F5F5}{ } 	 & 	 \cellcolor[HTML]{F5F5F5}{ } 	 & 	 88.2 	 & 	 \cellcolor[HTML]{F5F5F5}{ } 	 & 	 \cellcolor[HTML]{F5F5F5}{ } 	 & 	 \cellcolor[HTML]{F5F5F5}{ } 	 & 	 74.5 	 & 	 \cellcolor[HTML]{F5F5F5}{ } 	 & 	 \cellcolor[HTML]{F5F5F5}{ } 	 & 	 \cellcolor[HTML]{F5F5F5}{ } \\ 
MHRot \cite{DBLP:conf/nips/HendrycksMKS19} 	 & 	 77.6 	 & 	 \cellcolor[HTML]{F5F5F5}{ } 	 & 	 \cellcolor[HTML]{F5F5F5}{ } 	 & 	 \cellcolor[HTML]{F5F5F5}{ } 	 & 	 89.5 	 & 	 \cellcolor[HTML]{F5F5F5}{ } 	 & 	 \cellcolor[HTML]{F5F5F5}{ } 	 & 	 \cellcolor[HTML]{F5F5F5}{ } 	 & 	 83.6 	 & 	 \cellcolor[HTML]{F5F5F5}{ } 	 & 	 \cellcolor[HTML]{F5F5F5}{ } 	 & 	 \cellcolor[HTML]{F5F5F5}{ } \\ 
Reverse Distillation \cite{DBLP:conf/cvpr/DengL22} 	 & 	 - 	 & 	 \cellcolor[HTML]{F5F5F5}{ } 	 & 	 \cellcolor[HTML]{F5F5F5}{ } 	 & 	 \cellcolor[HTML]{F5F5F5}{ } 	 & 	 86.5 	 & 	 \cellcolor[HTML]{F5F5F5}{ } 	 & 	 \cellcolor[HTML]{F5F5F5}{ } 	 & 	 \cellcolor[HTML]{F5F5F5}{ } 	 & 	 - 	 & 	 \cellcolor[HTML]{F5F5F5}{ } 	 & 	 \cellcolor[HTML]{F5F5F5}{ } 	 & 	 \cellcolor[HTML]{F5F5F5}{ } \\ 
SSD \cite{DBLP:conf/iclr/SehwagCM21} 	 & 	 - 	 & 	 \cellcolor[HTML]{F5F5F5}{ } 	 & 	 \cellcolor[HTML]{F5F5F5}{ } 	 & 	 \cellcolor[HTML]{F5F5F5}{ } 	 & 	 90.0 	 & 	 \cellcolor[HTML]{F5F5F5}{ } 	 & 	 \cellcolor[HTML]{F5F5F5}{ } 	 & 	 \cellcolor[HTML]{F5F5F5}{ } 	 & 	 85.1 	 & 	 \cellcolor[HTML]{F5F5F5}{ } 	 & 	 \cellcolor[HTML]{F5F5F5}{ } 	 & 	 \cellcolor[HTML]{F5F5F5}{ } \\ 
CSI \cite{DBLP:conf/nips/TackMJS20} 	 & 	 52.4 	 & 	 \cellcolor[HTML]{F5F5F5}{ } 	 & 	 \cellcolor[HTML]{F5F5F5}{ } 	 & 	 \cellcolor[HTML]{F5F5F5}{ } 	 & 	 \underline{94.3} 	 & 	 \cellcolor[HTML]{F5F5F5}{ } 	 & 	 \cellcolor[HTML]{F5F5F5}{ } 	 & 	 \cellcolor[HTML]{F5F5F5}{ } 	 & 	 85.8 	 & 	 \cellcolor[HTML]{F5F5F5}{ } 	 & 	 \cellcolor[HTML]{F5F5F5}{ } 	 & 	 \cellcolor[HTML]{F5F5F5}{ } \\ 
\hline 
Supervised 	 & 	 \cellcolor[HTML]{F5F5F5}{ } 	 & 	 53.1 	 & 	 58.6 	 & 	 62.4 	 & 	 \cellcolor[HTML]{F5F5F5}{ } 	 & 	 55.6 	 & 	 63.5 	 & 	 67.7 	 & 	 \cellcolor[HTML]{F5F5F5}{ } 	 & 	 53.8 	 & 	 58.4 	 & 	 62.5 \\ 
SS-DGM \cite{DBLP:conf/nips/KingmaMRW14} 	 & 	 \cellcolor[HTML]{F5F5F5}{ } 	 & 	 - 	 & 	 - 	 & 	 - 	 & 	 \cellcolor[HTML]{F5F5F5}{ } 	 & 	 49.7 	 & 	 50.8 	 & 	 52.0 	 & 	 \cellcolor[HTML]{F5F5F5}{ } 	 & 	 - 	 & 	 - 	 & 	 - \\ 
Elsa \cite{DBLP:conf/bmvc/0001S0PC21} 	 & 	 \cellcolor[HTML]{F5F5F5}{ } 	 & 	 77.8 	 & 	 81.3 	 & 	 82.9 	 & 	 \cellcolor[HTML]{F5F5F5}{ } 	 & 	 80.0 	 & 	 85.7 	 & 	 87.1 	 & 	 \cellcolor[HTML]{F5F5F5}{ } 	 & 	 81.3 	 & 	 84.6 	 & 	 86.0 \\ 
\hline 
DeepSAD \cite{DBLP:conf/iclr/RuffVGBMMK20} 	 & 	 53.9 	 & 	 62.7 	 & 	 63.4 	 & 	 65.1 	 & 	 60.9 	 & 	 72.6 	 & 	 77.9 	 & 	 79.8 	 & 	 56.3 	 & 	 67.7 	 & 	 71.6 	 & 	 73.2 \\ 
DP VAE \cite{DBLP:journals/corr/abs-1911-04971} 	 & 	 61.7 	 & 	 65.4 	 & 	 67.2 	 & 	 69.6 	 & 	 52.7 	 & 	 74.5 	 & 	 79.1 	 & 	 81.1 	 & 	 56.7 	 & 	 68.5 	 & 	 73.4 	 & 	 75.8 \\ 
\hdashline 
\modelName{} (ours) 	 & 	 \textbf{\underline{81.4}} 	 & 	 \textbf{\underline{84.1}} 	 & 	 \textbf{\underline{85.3}} 	 & 	 \textbf{\underline{86.0}} 	 & 	 \textbf{91.5} 	 & 	 \textbf{92.5} 	 & 	 \textbf{\underline{97.1}} 	 & 	 \textbf{\underline{97.6}} 	 & 	 \textbf{\underline{86.1}} 	 & 	 \textbf{\underline{90.9}} 	 & 	 \textbf{\underline{92.3}} 	 & 	 \textbf{\underline{94.7}} \\ 
\tablebottom\end{tabular}
	}
	\caption{\label{tab:sota-sad-table}Comparison with SoTA methods over several datasets using the AUC in the one-vs-all protocol. The three blocks respectively contain one-class, semi-supervised and methods usable in both settings. \modelName{} is the best performing unified model. Underline indicates the overall best result, bold indicates the best semi-supervised method {\scriptsize (We re-evaluated Elsa, DP-VAE, SSAD, GOAD and ARNet on CIFAR100 and CUB-200)}.}
	\vspace{-6pt}
\end{table*}

\vspace{-5pt}
\section{Experiments}

\vspace{-1pt}
\subsection{Evaluation protocol}
\label{sec:ev-prot}

In the first \textbf{one-vs-all} protocol, one class is considered as normal and the others as anomalous. The final reported result is the mean of all runs obtained for each possible normal class. 
We consider various ratios $\gamma$ of anomaly data in the training dataset and for each, average the metrics on 10 random samples. The OCAD setting is a special case of the SSAD setting with $\gamma=0$. 
This protocol presents a realistic challenge and is widely used in AD literature. It shows the model ability to generalize to different anomalies without using a significant amount of training anomalies. Importantly, this test covers not only coarse object anomalies but also more subtle style and local anomalies where classes differ on minute details (as in CUB-200 and WMCA datasets)

The second protocol of \textbf{intra-dataset cross-type} is centered around face presentation attack detection (FPAD), which goal is to discriminate real faces from fake representations of faces. Training and test data are sampled from the same dataset, albeit with one tested attack type being unseen during training. We can thus evaluate the model generalization power and robustness to unseen attack types. Several attack types are considered: Paper Print (\textbf{PP}), Screen Recording (\textbf{SR}), Paper Mask (\textbf{PM}) and Flexible Mask (\textbf{FM}).

The final protocol of \textbf{OoD detection} consists in choosing an in-distribution (ID) dataset and an OoD one. After learning on the whole ID dataset, we evaluate how well the model discriminate ID samples from OoD samples. 

To evaluate the representation learning, we use the linear evaluation protocol which assesses the accuracy of a linear classifier trained on the encoder frozen representations.

Specifically, the considered datasets for the one-vs-all protocol are chosen to cover object and style anomalies:
\begin{itemize}[leftmargin=16pt,topsep=0pt,itemsep=0pt,partopsep=0pt,parsep=0pt]
\item \textbf{CIFAR-10} \cite{Krizhevsky2009LearningML}: an object recognition dataset composed of 10 wide classes with 6000 images per class.

\item \textbf{CIFAR-100} \cite{Krizhevsky2009LearningML}: more challenging version of CIFAR-10 with 100 classes, each containing 600 images.

\item \textbf{CUB-200} \cite{WelinderEtal2010}: a challenging fine-grained dataset consisting of 200 classes of birds, each containing around 50 images.
\end{itemize}

For the FPAD, we use the \textbf{WMCA} dataset \cite{DBLP:journals/tifs/GeorgeMGNAM20} containing more than 1900 RGB videos of real faces and presentation attacks. There are several types of attacks which cover object anomalies, style anomalies and local anomalies.

For the OoD detection, we add the commonly used datasets \textbf{SVHN} \cite{netzer2011} and \textbf{LSUN} \cite{yu2015}.

AD protocols report the area under the ROC curve (\textbf{AUROC}) averaged over all possible normal classes in the case of one-vs-all. When comparing two models with AUROCs $\alpha_1$ and $\alpha_2$, we use the relative error reduction $\scriptstyle \left((1{-}\alpha_2){-}(1{-}\alpha_1)\right)/(1{-}\alpha_1)$.

\subsection{Implementation details}
\noindent \textbf{Optimization.} Training is performed under SGD optimizer with nesterov momentum \cite{DBLP:conf/icml/SutskeverMDH13}, using a batch size of $B=1024$ and a cosine annealing learning rate scheduler \cite{DBLP:conf/iclr/LoshchilovH17} for both of the stages.\newline
\textbf{Data augmentation.} For the contrastive invariance transformations, we use random crop with rescale, horizontal symmetry, brightness jittering, contrast jittering, saturation jittering with gaussian blur and noise as in \cite{DBLP:conf/icml/ChenK0H20}. \newline
\textbf{Model design.} Regarding network architecture, we use a Resnet-50 \cite{DBLP:conf/cvpr/HeZRS16} ($\approx 25M$ parameters) for the backbone $f$. We consider two different memory scales: one after the third stage and another after the last stage. The associated memory layers are respectively of size $N^{(1)}_{\text{Mem}}=512$ and $N^{(2)}_{\text{Mem}}=256$ with an inverse temperature $\beta=2$, along with a pattern sampling ratio of $r^{(1)}=0.3$ and $r^{(2)}=1$. The choices of memory size and sampling ratios are respectively discussed in \cref{sec:mem-size} and \cref{sec:sampling-ratio}. The scale confidence factors are set to increase exponentially as $\lambda^{(\scaleK)}=2^{\scaleK-1}$, and the variance loss factor is fixed to $\lambda_V=0.05$ after optimization on CIFAR-10. We use $\tau=0.1$ as suggested in \cite{DBLP:conf/icml/ChenK0H20}. For the anomaly distance loss, we choose a margin size of $M=2$. 

\subsection{Comparison to the state-of-the-art}

\subsubsection{One-vs-all}
\label{sec:sota-one-vs-all}

This section compares \modelName{} with SoTA AD methods on the \textbf{one-vs-all protocol}, in the one-class setting and the semi-supervised setting when possible. 

Considered one-class methods are hybrid models \cite{DBLP:conf/nips/ScholkopfWSSP99}, reconstruction error generative model \cite{DBLP:conf/acpr/TuluptcevaBFK19}, the knowledge distillation method \cite{DBLP:conf/cvpr/DengL22}, pretext tasks methods \cite{DBLP:conf/iclr/BergmanH20,DBLP:conf/nips/HendrycksMKS19}, and the two-stage method \cite{DBLP:conf/nips/TackMJS20}. We also consider semi-supervised methods such as density estimation methods \cite{DBLP:conf/nips/KingmaMRW14}, and two-stage AD \cite{DBLP:conf/bmvc/0001S0PC21}. To further show the disadvantages of classical binary classification, we also include a classical deep classifier trained with batch balancing between normal samples and anomalies. Lastly, unified methods usable in both one-class and semi-supervised learning are included with the reconstruction error model \cite{DBLP:journals/corr/abs-1911-04971}, and direct anomaly distance models \cite{DBLP:conf/iclr/RuffVGBMMK20}. For a fair comparison in the same conditions, we take the existing implementations or re-implement and evaluate ourselves all one-class methods, except \cite{fei2020attribute, DBLP:conf/iclr/BergmanH20, DBLP:conf/nips/TackMJS20}. The results are presented in \cref{tab:sota-sad-table}.

First of all, we can notice the classical supervised approach falls far behind anomaly detectors on all datasets. This highlights the importance of specialized AD models, as classical models are likely to overfit on anomalies.

Furthermore, \modelName{} overall performs significantly better than all considered detectors on various datasets with up to 62\% relative error improvement on CIFAR-10 and $\gamma=0.01$. Although performance greatly increases with more anomalous data, it remains highly competitive with only normal samples. In the OC setting, \modelName{} outperforms all methods specialized for OC including pretext task methods, reconstruction error methods and by far hybrid methods. The usage of memory in \modelName{} is much more efficient than the memory for reconstruction used in MemAE. Indeed, while we learn the memory through contrastive learning, MemAE and others \cite{DBLP:conf/iccv/GongLLSMVH19,DBLP:conf/cvpr/ParkNH20} learned it via the pixel-wise reconstruction loss. Their normal prototypes are much more constrained and therefore less semantically rich and generalizable. For SSAD, \modelName{} reduces SoTA error gap on nearly all anomalous data ratio. Its multi-scale anomaly detectors allow capturing more fine-grained anomalies as we can see in the CUB-200 results. 

Finally, \modelName{} performs very well in both OC and SSAD while other unified methods generally fail in the OC setting. In this regard, \modelName{} is to the best of our knowledge the \textit{first efficient unified anomaly detector}. We also note that the change from OCAD to SSAD in our model was done with minimal hyperparameter tuning. This is due to the first training step being shared between OC and SS settings.
\vspace{-0.5cm}
\subsubsection{Out-of-Distribution (OoD) detection}
In \cref{tab:sota-ood}, we compare \modelName{} to SoTA models with CIFAR-10 as ID dataset and OoD datasets SVHN, LSUN and CIFAR-100. Our model keeps among the best performance with a relative error improvement of 37\% with the second best performing method on the challenging CIFAR-10 vs CIFAR-100. \modelName{} performs similarly well as the SoTA baseline CSI \cite{DBLP:conf/nips/TackMJS20}, however the latter mainly relies during inference on keeping in memory the entire training set features. This results in a very high memory footprint on huge training sets, while our model ingeniously learns a fixed size memory. Moreover, CSI often performs poorly on datasets with a few training normal samples as can be seen on one-vs-all CUB-200 (50 images per class) results. 

It is also worth noting that algorithms in \cite{DBLP:journals/corr/abs-2106-03844,RethingkingOoD2023} obtain higher performance on this task, however, they rely on models pretrained on an additional large-scale dataset such as ImageNet-21K \cite{imagenet21k}. Therefore, comparing directly those methods to \modelName{}, being trained from scratch on a smaller dataset, is not fair as there is an overlap in nature of the pre-training samples used in \cite{RethingkingOoD2023} and anomalous samples to be detected.

Finally, we note that in this protocol the normal class is composed of several sub-classes. This shows the ability of our memory modules to cover multi-modal normal cues.
\begin{table}[tb]
	\centering
	\resizebox{0.76\columnwidth}{!}{%
		\begin{tabular}{l|ccc} 
\tabletop\textbf{AUROC (\%)} 	 & 	 \multicolumn{3}{c}{\textbf{CIFAR-10 $\rightarrow$}} \\ 
\textbf{Models / OoD data} 	 & 	 \textbf{SVHN} 	 & 	 \textbf{LSUN} 	 & 	 \textbf{CIFAR-100} \\ 
\hline 
GOAD  \cite{DBLP:conf/iclr/BergmanH20}	 & 	 96.3 	 & 	 89.3 	 & 	 77.2 \\ 
MHRot \cite{DBLP:conf/nips/HendrycksMKS19} 		 & 	 97.8 	 & 	 92.8 	 & 	 82.3 \\ 
CSI \cite{DBLP:conf/nips/TackMJS20}	 & 	 \textbf{99.8} 	 & 	 97.5 	 & 	 89.2 \\ 
\hdashline 
\modelName{} (ours) 	 & 	 99.2 	 & 	 \textbf{97.5} 	 & 	 \textbf{89.2} \\ 
\tablebottom\end{tabular}
	}
	\caption{Comparison of \modelName{} with SoTA methods on OoD detection protocol with CIFAR-10 as ID dataset.}
	\label{tab:sota-ood}
\end{table}

\subsubsection{Face presentation attack detection (FPAD)}

\cref{tab:sota-fpad-table} compares our model on the FPAD intra-dataset cross-type protocol with SoTA methods presented in \cref{sec:sota-one-vs-all}. 

Without any further tuning for face data, our method improves FPAD performance on WMCA with an error relative improvements of up to 14\% on paper prints. It outperforms existing anomaly detectors on all unseen attack type, including the OC setting. We can also notice that it reduces the error gap between coarse attacks (PM, FM) and harder fine-grained attacks (PP, SR) thanks to its multi-scale AD.

\begin{table}[h]
	\centering
	\resizebox{0.75\columnwidth}{!}{%
		\begin{tabular}{l|ccccc} 
\tabletop\textbf{AUROC (\%)} 	 & 	 \multicolumn{5}{c}{\textbf{WMCA}} \\ 
\textbf{Models / Kind} 	 & 	 \textbf{All} 	 & 	 \textbf{PP} 	 & 	 \textbf{SR} 	 & 	 \textbf{PM} 	 & 	 \textbf{FM} \\ 
\hline 
PIAD  \cite{DBLP:conf/acpr/TuluptcevaBFK19}	 & 	 76.4 	 & 	 \cellcolor[HTML]{F5F5F5}{ } 	 & 	 \cellcolor[HTML]{F5F5F5}{ } 	 & 	 \cellcolor[HTML]{F5F5F5}{ } 	 & 	 \cellcolor[HTML]{F5F5F5}{ } \\ 
GOAD \cite{DBLP:conf/iclr/BergmanH20} 	 & 	 86.1 	 & 	 \cellcolor[HTML]{F5F5F5}{ } 	 & 	 \cellcolor[HTML]{F5F5F5}{ } 	 & 	 \cellcolor[HTML]{F5F5F5}{ } 	 & 	 \cellcolor[HTML]{F5F5F5}{ } \\ 
MHRot \cite{DBLP:conf/nips/HendrycksMKS19} 	 & 	 81.3 	 & 	 \cellcolor[HTML]{F5F5F5}{ } 	 & 	 \cellcolor[HTML]{F5F5F5}{ } 	 & 	 \cellcolor[HTML]{F5F5F5}{ } 	 & 	 \cellcolor[HTML]{F5F5F5}{ } \\ 
PuzzleGeom \cite{DBLP:conf/avss/JezequelVBH21}	 & 	 85.6 	 & 	 \cellcolor[HTML]{F5F5F5}{ } 	 & 	 \cellcolor[HTML]{F5F5F5}{ } 	 & 	 \cellcolor[HTML]{F5F5F5}{ } 	 & 	 \cellcolor[HTML]{F5F5F5}{ } \\ 
\hline 
Supervised 	 & 	 \cellcolor[HTML]{F5F5F5}{ } 	 & 	 78.3 	 & 	 77.1 	 & 	 80.7 	 & 	 81.9 \\ 
Elsa \cite{DBLP:conf/bmvc/0001S0PC21} 	 & 	 \cellcolor[HTML]{F5F5F5}{ } 	 & 	 86.1 	 & 	 84.3 	 & 	 89.2 	 & 	 89.1 \\ 
SadCLR \cite{DBLP:conf/icpr/Jezequel22CR}	 & 	 \cellcolor[HTML]{F5F5F5}{ } 	 & 	 89.8 	 & 	 88.5 	 & 	 92.7 	 & 	 91.9 \\ 
\hline 
DP VAE \cite{DBLP:journals/corr/abs-1911-04971} 	 & 	 53.9 	 & 	 - 	 & 	 - 	 & 	 - 	 & 	 - \\ 
DeepSAD \cite{DBLP:conf/iclr/RuffVGBMMK20} 	 & 	 71.2 	 & 	 79.9 	 & 	 80.3 	 & 	 81.8 	 & 	 83.4 \\ 
\hdashline 
\modelName{} (ours) 	 & 	 \textbf{86.9} 	 & 	 \textbf{91.3} 	 & 	 \textbf{89.8} 	 & 	 \textbf{93.0} 	 & 	 \textbf{92.7} \\ 
\tablebottom\end{tabular}
	}
	\caption{Comparison of \modelName{} with SoTA methods over face anti-spoofing datasets in the cross-type protocol with regards to AUC. The columns indicate the type of presentation attack unseen during training. Bold indicates the best.}
	\label{tab:sota-fpad-table}
\end{table}

\subsection{Ablation study}
\label{sec:abl-study}

In this section we study the impact of the multi-scale memory layers in the two training stages and show they are essential to our model performance.

First, we evaluate using linear evaluation on CIFAR-10 and CIFAR-100 how the memory affects the contrastive learning of the encoder representations. As we can see in \cref{tab:abl}(a), the inclusion of the memory layers on the first branch drastically improves the quality of the encoder representations. We hypothesis that, as shown in \cite{DBLP:conf/iccv/DwibediATSZ21}, the inclusion of prototypical samples in one of the branch allows to contrast positive images against representative negatives. This alleviates the need for large batch size, and highly reduces the multi-scale contrastive learning memory usage.

To support the importance of memory during AD, we compare the performance of the anomaly detector with k-means centroids or with the normal prototypes learned during the first stage. In the first case, we train the anomaly detectors with the same procedure but instead of fetching the Hopfield layer output we use the closest k-means centroid. The results on CIFAR datasets are presented in \cref{tab:abl}(b). 

Lastly we measure in \cref{tab:abl}(a)(b) the impact of multi-scale AD by comparing the single-scale model only using the last feature map, and our two-scale model. While we sacrifice some of the memory and training time for the additional scale, our AD performance is improved significantly with up to 10\% error relative improvement on CIFAR-10.

\begin{table}[tb]
	\centering
	\resizebox{0.76\columnwidth}{!}{%
		\begin{tabular}{cc|cc}
			\tabletop
			\textbf{Memory} & \textbf{Multi-scale} & \textbf{CIFAR-10}                                & \textbf{CIFAR-100}                                \\ 
			\hline
			\multicolumn{4}{c}{(a) Representation Learning (Linear Evaluation)}                                                                                       \\ 
			\hline
			-               & -                    & 88.2 \phantom{\metricrel{+0.0}}                            & 80.5 \phantom{\metricrel{+0.0}}                  \\
			$\checkmark$    & -                    & 91.4 \metricrel{+3.2}          & 84.1 \metricrel{+3.6}           \\
			$\checkmark$    & $\checkmark$         & \textbf{91.9} \metricrel{+0.5} & \textbf{84.8} \metricrel{+0.7}  \\ 
			\hline
			\multicolumn{4}{c}{(b) Anomaly Detection (One-vs-all)}                                                                                      \\ 
			\hline
			-               & -                    & 88.1 \phantom{\metricrel{+0.0}}                                            & 82.7  \phantom{\metricrel{+0.0}}                \\
			$\checkmark$    & -                    & 90.5 \metricrel{+2.4}          & 85.3 \metricrel{+2.6}           \\
			$\checkmark$    & $\checkmark$         & \textbf{91.5} \metricrel{+1.0} & \textbf{86.1} \metricrel{+0.8} \\
			\tablebottom
		\end{tabular}
	}
	\caption{\label{tab:abl}Ablation study of our model during the two stages. (a) We perform a linear evaluation of representation learning, (b) we assess the one-vs-all anomaly detection AUC. The baseline is the model using only the last feature map, with kmeans cluster centroids as its normal prototypes.}
\end{table}

\subsection{Memory size}
\label{sec:mem-size}

The memory size must be carefully chosen at every scale to reach a good balance between the normal class prototype coverage and the memory usage during training and inference. This section presents some rules of thumb regarding the sizing of memory layers depending on the scale. 

\begin{figure}[tb]
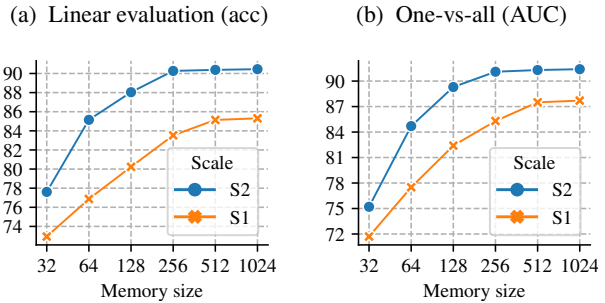

	\centering
	\begin{subfigure}[b]{0.5\columnwidth}
		\centering
		\caption{\label{fig:mem-size} Linear evaluation (acc)}
		\resizebox{0.9\columnwidth}{!}{
			\begin{adjustbox}{clip,trim=0.1in 0.1in 0.11in 0}
				\input{figures/plot_exp_mem1.pgf}
			\end{adjustbox}
		}
	\end{subfigure}%
	~ 
	\begin{subfigure}[b]{0.5\columnwidth}
		\centering
		\caption{\label{fig:mem-scale} One-vs-all (AUC)}
		\resizebox{0.9\columnwidth}{!}{
			\begin{adjustbox}{clip,trim=0.1in 0.1in 0.11in 0}
				\input{figures/plot_exp_mem2.pgf}
			\end{adjustbox}
		}
	\end{subfigure}%
	\vspace{-3pt}
	\caption{Memory experiments on CIFAR-10.}
\end{figure}

We start by plotting the relation between the last scale memory size, the representation separability, and the final AD performance in \cref{fig:mem-size}. As one could expect, higher memory size produces better quality representations and more accurate anomaly detector. However, we can notice that increasing the memory size above 256 has significantly less impact on the anomaly detector performance. Therefore a good trade-off between memory usage and performance seems to be at 256 on CIFAR-10. \newline
Furthermore, we study the impact of feature map scale on the required memory size. In \cref{fig:mem-scale} we fix the last scale memory size to 256 and look at the AD accuracy for different ratios of memory size between each scale. Interestingly, lower scales seem to benefit more from a larger memory than higher scales. We can hypothesis that the feature vectors of more local texture-oriented features will be richer and more variable than global object-oriented ones. Memory layers thus need to be of higher capacity to capture the complexity.

\subsection{Spatial sampling ratios}
\label{sec:sampling-ratio}

Sampling ratios $r$ are introduced during the first step in order to reduce the amount of patterns considered in the contrastive loss, and consequently the similarity matrix size. In low scales, we can expect nearby samples to be quite similar. Therefore, it is not as detrimental to the training to skip some of the available patterns.  

\begin{figure}[tbh!]
	\centering
	\label{fig:sampling-ratio}
	\begin{adjustbox}{clip,trim=0 0.05in 0 0.05in}
	\resizebox{0.8\columnwidth}{!}{
	\input{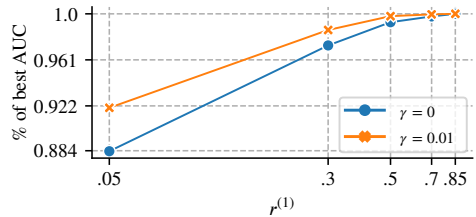}
	}
	\end{adjustbox}
	\vspace{-10pt}
	\caption{Sampling ratio experiments on CIFAR-10.}
\end{figure}

To guide our choice of sampling ratio, we measure our anomaly detection AUC with various sampling ratio and anomalous data ratio $\gamma$. Since the last scale feature maps are spatially very small, we only vary the first scale ratio $r^{(1)}$ and set $r^{(2)}=1$. The batch size if fixed throughout the experiments. The results are displayed in \cref{fig:sampling-ratio}. We can see that low sampling ratios ($\gamma<0.3$) significantly decrease the AD performance. However the gain in performance for higher ratios is generally not worth the additional computational cost: by more than doubling the amount of sampled patterns, we only increase the relative AUC by 2\%. 

\vspace{-7pt}
\section{Conclusion and future work}\vspace{-3pt}
In this paper, we present a new two-stage anomaly detection model which memorizes normal class prototypes in order to compute an anomaly deviation score. By introducing normal memory layers in a contrastive learning setting, we can first jointly learn the encoder representations and a set of normal prototypes. This improves the quality of the learned representations, and allows a strong memorization of normal samples. The normal prototypes are then used to train a simple detector in a unified framework for one-class or semi-supervised AD. Furthermore, we extend these prototypes to several scales making our model more robust to different anomaly sizes. Finally, we assess its performance on a wide array of dataset containing object, style and local anomalies. \modelName{} greatly outperforms SoTA performance on all datasets and different anomalous data regimes.

For future work, we could explore the use of multi-scale anomaly score for anomaly localization. Indeed, in the one-class setting of our model we could merge the several scale anomaly maps into a single heatmap.

{\small
\bibliographystyle{ieee_fullname}
\bibliography{refs_final}
}

\end{document}